\documentclass[conference]{IEEEtran}
\IEEEoverridecommandlockouts
\usepackage{cite}
\usepackage{amsmath,amssymb,amsfonts}
\usepackage{algorithmic}
\usepackage{graphicx}
\usepackage{textcomp}
\usepackage[table]{xcolor}

\usepackage{hyperref}

\usepackage{booktabs}
\usepackage{multirow}

\usepackage{xspace}

\usepackage{enumitem}

\usepackage[skins,listings]{tcolorbox}
\def\BibTeX{{\rm B\kern-.05em{\sc i\kern-.025em b}\kern-.08em
    T\kern-.1667em\lower.7ex\hbox{E}\kern-.125emX}}
    
\newboolean{showcomments}
\setboolean{showcomments}{true}
\ifthenelse{\boolean{showcomments}}
{\newcommand{\nb}[2] {
\fcolorbox{black}{gray!20}{\bfseries\sffamily\scriptsize#1:}
{\sf\small$\blacktriangleright$\textit{#2}$\blacktriangleleft$}
}
}
{\newcommand{\nb}[2]{}
}

\usepackage{enumitem}
\usepackage{ifthen}
\usepackage{soul}   
    
\usepackage[frozencache=true,cachedir=.]{minted}

\newcommand{\etal}{\textit{et al.}\xspace}
\newcommand{\tool}{\textsc{uncertainty-wizard}\xspace} %
\newcommand{\toolUpper}{\textsc{Uncertainty-Wizard}\xspace} %
\newcommand{\tfkeras}{\textsc{tf.keras}\xspace} %

\newcommand\exampleReferences[1]{\hyperref[lst:ex_stochastic]{{\color{green!50!black}{(\textsc{#1})}}}} %
\newcommand\exampleReferencesNoBrackets[1]{{\color{green!50!black}{\textsc{#1}}}} %

\begin{document}
\pagenumbering{arabic} 
\pagestyle{plain}

\title{\toolUpper: Fast and User-Friendly Neural Network Uncertainty Quantification\\
\thanks{
This work was partially supported by the H2020 project PRECRIME,
funded under the ERC Advanced Grant 2017 Program (ERC Grant Agreement n. 787703).\newline
Accepted at ICST2021. 
© 2021 IEEE. Personal use of this material is permitted. Permission from IEEE must be
obtained for all other uses, in any current or future media, including
reprinting/republishing this material for advertising or promotional purposes, creating new
collective works, for resale or redistribution to servers or lists, or reuse of any copyrighted
component of this work in other works.
}
}

\author{\IEEEauthorblockN{Michael Weiss and
Paolo Tonella}
\IEEEauthorblockA{
Universit\`a della Svizzera italiana, \\
Lugano, Switzerland\\
\{michael.weiss ¦ paolo.tonella\}@usi.ch}}
\maketitle


\begin{abstract}
    Uncertainty and confidence have been shown to be useful metrics in a wide variety of techniques proposed for deep learning testing, including test data selection and system supervision.
    We present \tool, a tool that allows to quantify such uncertainty and confidence in  artificial neural networks.
    It is built on top of the industry-leading \tfkeras deep learning API and it provides a near-transparent and  easy to understand interface.
    At the same time, it includes major performance optimizations that we benchmarked on two different machines and different configurations.
\end{abstract}


\section{Introduction}
\label{sec:introduction}

Modern software systems increasingly include artificial neural networks (ANNs), a powerful  machine learning technique used for processing and interpreting large amounts of data even on resource constrained devices.
Popular machine learning frameworks interfaces, such as Tensorflow's \tfkeras offer a high level of abstraction on complex statistical processes and thus allow even software engineers without extensive experience in machine learning to create well working ANNs.

However, the integration of such statistical components into a software system comes with major challenges to software testing:
Given the stochastic nature of ANNs and the fact that their input spaces are too big to be tested exhaustively and are often intrinsically ambiguous, any code that contains ANNs should always expect that incorrect predictions may occasionally happen. 
An important technique to deal with this is \emph{uncertainty} -- or \emph{confidence} -- quantification, 
i.e., the calculation of a score which measures the probability or severity of wrong predictions (uncertainty) or the probability of a correct prediction (confidence)\footnote{As uncertainty and confidence quantification are perfect complements, we will, w.l.o.g., refer only to uncertainty quantification for the remainder of the paper.}.

The machine learning literature provides a range of different uncertainty quantification techniques. 
However, implementing them is often not as easily achieved as creating a standard model, and requires in-depth study of the related literature. 
As part of our recent work on the empirical evaluation of uncertainty quantification \cite{Weiss2021}, we have developed and released \tool,
a tool that allows to quantify confidence and uncertainty of \tfkeras models using a simple interface, associated with a performance-optimized implementation. 
With our tool the goal is to make uncertainty quantification equally easy and practical to use as the creation of a regular \tfkeras model.
In this paper, we present our design choices, challenges and evaluations towards reaching this goal. 
Our software, installation instructions and documentation are available at:
\begin{center}
    \href{https://github.com/testingautomated-usi/uncertainty-wizard}{github.com/testingautomated-usi/uncertainty-wizard}.
\end{center}

\section{Applications In Software Testing}
\label{sec:applications}

Software testing researchers have proposed and used a range of techniques to detect inputs on which the ANN under test has high probability of misprediction. These include 
surprise adequacy\cite{Kim2018}, dissector \cite{Wang2020} and autoencoders \cite{Stocco2020}.
Kim et al.\cite{Kim2018} introduce two measures of the degree of surprise of a new input w.r.t. the training set. Wang et al.\cite{Wang2020} combine multiple predictions from ``dissected'' ANN models into a validity score for the given input. Stocco et al.\cite{Stocco2020} measure the reconstruction error of an autoencoder to estimate the probability that the given input may lead a self-driving car to crash.
All these approaches rely on the  detection of inputs which were underrepresented during training and thus have a high probability of failure.
The related metrics can thus be regarded as uncertainty quantification scores.

Only a few papers in the testing literature take advantage of well established uncertainty quantification from the machine learning literature, such as \emph{MC-Dropout} or \emph{Deep Ensembles} (notable exceptions are, e.g., the works by Zhang et al.\cite{Zhang2020} and by Berend et al.\cite{Berend2020}). Both  techniques are very popular and well researched in machine learning, and they can be applied to almost every classical ANN architecture. They also have interesting theoretical properties, which go beyond the capability of detecting underrepresented data.
At least partly, this may be due to the challenges faced when implementing these approaches, often available only through their mathematical formulation. This motivated the creation of \tool,
which makes such techniques usable by means of a simple, near-transparent interface.
\section{Implemented Approaches}
\label{sec:approaches}

Aiming for an easy and highly compatible adoption in software testing, we focus on approaches which have no or only basic requirements about the ANN's architecture.
For an easy start into the literature about uncertainty and its quantification, the reader can refer to our empirical study of uncertainty approaches \cite{Weiss2021} and the tutorial by Jospin \etal \cite{Jospin2020}.

\paragraph{Monte-Carlo Dropout (MC-Dropout)\cite{Gal2016}} 
This technique quantifies uncertainty by sampling outputs of multiple non-deterministic forward passes in an ANN,
made to infer an output distribution. 
This output distribution then allows to choose the most likely prediction and to quantify its uncertainty.
To achieve  randomness in the forward passes, MC-Dropout profits from the fact that most classical ANNs use \textit{dropout layers} as regularization technique during training:
These layers independently drop every activation which is passed through them with a predefined probability 
$P_{drop} \in [0,1]$.
In MC-Dropout, the same is done at prediction time, to collect multiple, non-deterministic samples of the output.
In \tfkeras, enabling such sampling would require major code changes, amongst other reasons due to the high abstraction level of the popular and simple \emph{Sequential API}. 

\paragraph{Deep Ensembles\cite{Lakshminarayanan2017}}
Deep Ensembles are a collection of $n > 1$, $ n \in \mathbb{N} $ independent \emph{atomic models}. 
They typically share the same architecture, but due to different (randomly set) initial weights and random influences during training (e.g. data augmentation or regularization techniques) they complete their training in slightly different states.
Similar to MC-Dropout, this  allows to collect  $n$ ANN outputs for the same input and to use them to infer a predictive distribution.
It is worth noting that Deep Ensembles were  found to be, on average but not strictly, the most effective  uncertainty quantification approach\cite{Weiss2021, Ovadia2019}. 
The biggest disadvantage of Deep Ensembles is their high computational requirements, as every model execution (e.g. training, prediction) has to be done $n$ times.
Storing atomic models on the file system when not in use, and parallel execution of multiple atomic models can improve the runtime and memory usage, but requires the writing of a large amount of boilerplate code.
Parallel and thus fast processing of deep ensembles, without the need for boilerplate code, is a main contribution of \tool.

\paragraph{Point Predictors} 
Point Predictions refer to predictions which are calculated based on a single ANN forwards pass (i.e., a non-sampled single prediction).
In particular, in classification problems with a softmax output layer, a single point prediction still allows to quantify uncertainty based on the model's output.
For example, the class with the highest softmax value is chosen as predicted class and the softmax value of that class is  used as confidence. 
Other approaches compare the highest and the second highest predictions\cite{Zhang2020}; others compute the entropy of the softmax outputs\cite{Weiss2021}.
Point predictors are usually simple to implement and fast to compute, but theoretically not well grounded, which may bring major practical shortcomings\cite{Gal2016}. 
Nonetheless, there are cases where they outperform MC-Dropout and Deep Ensembles in detecting wrong ANN predictions\cite{Weiss2021}.
We include them in \tool, primarily to allow easy comparison, as well as for use in extremely resource-constrained environments. 

What these three approaches have in common is that they all infer: (1) a final prediction and (2) an uncertainty (or confidence) value.
We call the functions responsible for such task \emph{quantifiers}. 
We can distinguish between two coarse-grain categories of quantifiers: \emph{sampling-based quantifiers (SBQ)}, used by MC-Dropout and Deep Ensembles, and \emph{Point-Predictor quantifiers (PPQ)}, which define quantifiers based on single-pass and single-model ANN outputs.

\section{Features and Interface}
\label{sec:interface}

This section provides an overview of the features implemented in \tool and the API exposed to users.
Basic usage examples are given in Listings \ref{lst:ex_stochastic} (\emph{Stochastic Model}, which combines the use of MC-Dropout and Point Predictions) 
and \ref{lst:ex_lazy_ensemble} (Deep Ensemble). 
For the remainer of this paper, we will refer to these examples using the notations \exampleReferencesNoBrackets{S$n$} (for the stochastic model) 
and \exampleReferencesNoBrackets{E$n$} (for the deep ensemble), where $n$ denotes a line number.

The following list provides an overview of the most important features and   APIs implemented in \tool.

\newcommand\litem[1]{\item{\bfseries #1\\}}

\begin{enumerate}
\litem{Out of the Box Uncertainty: \lstinline{predict_quantified}} 
As a key feature, every model created using our tool exposes a \lstinline{predict_quantified} function which allows to make ANN predictions and quantify their uncertainty:
Passing a suitable quantifier, \lstinline{predict_quantified} collects the ANN outputs and uses the quantifier to calculate predictions and uncertainties. \exampleReferences{S10, S15, E16}
 
\end{enumerate}
The following features are specific to MC-Dropout and Point-Predictors:
\begin{enumerate}[resume]
\litem{Support for Sequential and Functional Models} 
\tfkeras provides multiple APIs to implement ANNs.
The simplest one is the Sequential API, where layer instances are stacked on top of each other. 
The actual function calls between these layers is then transparently inferred by \tfkeras.
The only way to enable Dropout during predictions requires to set a specific argument in the function calls between the layers - which thus cannot easily be done using the \tfkeras Sequential API. 
Hence, the \tfkeras sequential API cannot be used to perform MC-Dropout.
\tool in turn allows to easily create sequential models for MC Dropout by only changing one line of code: 
Replacing the \tfkeras constructor \lstinline{Sequential()} with the corresponding \tool's constructor \lstinline{StochasticSequential()}. This automatically configures any later added Dropout (or other type of randomized) layers to be enabled when calling \lstinline{predict_quantified} with a sampling-based quantifier \exampleReferences{S2}.

\litem{Point-Predictor and MC-Dropout in one model} 
To use MC Dropout in plain \tfkeras, the user has to enable dropout when creating the model instance and it is hard to change this later.
This makes it also hard to apply both PPQ and SBQ on the same model instance, which is only possible by duplicating the model: One will have dropout enabled and the other dropout disabled. 
Provided that both  models share the same weights, such duplication is inefficient and not user friendly.
\tool allows to use a single model instance for both PPQ and SBQ. 
Based on the type of the passed quantifiers and in a fully transparent way, \lstinline{predict_quantified} dynamically enables and disables randomization. Accordingly, whether to collect ANN outputs from a single forward pass or to collect multiple samples is dynamically decided as well, depending on the passed quantifiers \exampleReferences{S10, S15}.

\litem{Create from pre-trained, regular \tfkeras models} 
\tool allows to create new models from regular \tfkeras models 
by calling \lstinline{uwiz.models.stochastic_from_keras(<keras_model>)}.
Randomized layers such as Dropout are automatically recognized and prepared for use with SBQs.
This is particularly useful when working with pretrained models, 
which were trained without the use of \tool.

\end{enumerate}

The following features are specific to Deep Ensembles:

\begin{enumerate}[resume]
\litem{Transparent Laziness} 
Holding all atomic models of a Deep Ensemble in GPU or system memory can easily exceed the available capacities.
To provide a scalable implementation, \tool  handles models lazily:
Our \lstinline{LazyEnsemble} objects do not actually keep any atomic model in memory, but instead it
persists them on the file system and automatically loads them when used \exampleReferences{E9}.
To run a task on all atomic models, users can submit tasks to the ensemble,
in a functional way using \lstinline{create(<supplier>, ...)}, \lstinline{model.modify(<mapping>, ...)} or \lstinline{model.consume(<consumer>, ...)}, where the passed functions expect an atomic model as input (except for \lstinline{create}), and return an atomic model as output (except for \lstinline{consume}).
All tasks can also return a generic $T$ which is an arbitrary value which will be collected in a list from all atomic models and returned from the called ensemble function, i.e., of \lstinline{create}, \lstinline{modify} or \lstinline{consume}. 
An example of this is given in \exampleReferencesNoBrackets{E7}, where a \lstinline{supplier} that trains a model returns the training history.
The collected training histories are then returned in \exampleReferencesNoBrackets{E13}.
We decided to support a generic result $T$ to be returned and collected from all atomic models to provide developers with a quantification mechanism that can potentially go beyond uncertainty estimation, since any arbitrary distribution of metrics can be computed in this way. 
Note that, for the purpose of uncertainty quantification the returned $T$ value of a \lstinline{consumer} is typically the model output.
Hence, for uncertainty quantification, we provide also the following two wrappers of \lstinline{consume}, which allow to pass quantifiers and to infer prediction and uncertainties from the atomic models' outputs:
\begin{itemize}
    \item \lstinline{predict_quantified(x, quantifier, ...)}, where \tool internally creates a consumer to calculate the model outputs for numpy inputs \lstinline{x}.
    \item \lstinline{quantify_predictions(quantifier, consumer, ...)}, where the generic results of the consumer are expected to be the models' outputs, while inputs are loaded within the consumer.
\end{itemize}

\litem{Various Parallelization Contexts} 
If a system's hardware is powerful enough that sequential processing (e.g. training) of the atomic models does not need to use the full system capacity, there is a clear potential for performance improvements by parallel processing. 
\tool's deep ensemble methods can be parallelized by simply providing the number of desired processes as an additional parameter \exampleReferences{E14, E18}.
 
\litem{Plain \tfkeras models} 
The above described \lstinline{create}, \lstinline{modify} and \lstinline{consume} are not uncertainty quantification specific, and the models used in these function are plain \tfkeras models.
Hence, applications of our \lstinline{LazyEnsemble} implementation may go beyond uncertainty quantification: 
In ANN testing, often a procedure has to be performed multiple times to gain statistical significance or on multiple models, e.g., during unit (model) or mutation testing.
\exampleReferences{E3}
\end{enumerate}

Further utilities in our APIs include:
\begin{enumerate}[resume]
\litem{Multi-quantifier support} 
Both academic as well as practical use-cases may require the execution of multiple quantifiers at the same time\cite{Weiss2021}.
Thus \lstinline{predict_quantified}  accepts a list of quantifiers in place of a single one. 
When called with such a list of quantifiers, \tfkeras will return a corresponding list of prediction and uncertainty tuples \exampleReferences{S17}.
\litem{Uncertainty-Confidence conversions} 
\lstinline{predict_quantified} and \lstinline{quantify_predictions} allow  passing  an optional \lstinline{as_confidence} flag. If set to \lstinline{True}, the calculated uncertainties are converted into confidences. If set to \lstinline{False}, the calculated confidences are converted to uncertainties. With the default \lstinline{None}, no conversions are performed.
\litem{Full access to all \tfkeras functionality} 
All our models are based on regular \tfkeras models,
which can be transparently used as such.
Hence, \tool does not limit the user's ability to use \tfkeras functions when our library is applied to existing \tfkeras models
\exampleReferences{S19,E3}.
\litem{High configurability} 
The flexibility of \tool  exceeds the basic functionalities explained in this section. 
By design, \tool accepts the injection of configurations and processes replacing our defaults.
Examples include (1) the ability to create custom quantifiers,
(2) functions to allocate specific number of processes in lazy ensembles to specific devices (GPUs or CPU),
(3) the ability to create custom stochastic layers in stochastic models, and
(4) the ability to override a model's save and load processes in lazy ensembles.
\end{enumerate}

\begin{listing}[t]
    \begin{minted}[fontsize=\scriptsize, linenos]{python}
# Create MC-Dropout capable model
model = uwiz.models.StochasticSequential()
# Use as a regular tf.keras model
model.add(keras.layers.Dense(100)
model.add(keras.layers.Dropout(0.2)
model.add(keras.layers.Softmax(10)
model.compile( ... )
model.fit( ... )
# Predict as Point-Predictor w/ confidence
pred_pp, pcs = model.predict_quantified(
    x_test, quantifier='pcs')
# Predict w/ MC-Dropout w/ uncertainty;
#   results (list) contains two (pred, unc) 
#   tuples, one per quantifier
results = model.predict_quantified(
    x_test, num_samples=100, 
    quantifier=['pred_entropy', 'var_ratio'])
# Get as plain (non-uwiz) keras model
keras_model = model.inner

    \end{minted}
    \caption{Stochastic Sequential Model}
    \label{lst:ex_stochastic}
\end{listing}

\section{Technical Challenges and Implementation}
\label{sec:implementation}

Our overall design goals were: (1) to implement an end-to-end tool, which supports all the steps required to perform uncertainty quantification; (2) to keep the performance impacts as little as possible; and, (3) to build whenever possible on (most likely stable) high-level \emph{Tensorflow} methods, in order to reduce the possible compatibility issues with future versions of \emph{Tensorflow}.

\subsection{MC-Dropout and Point-Predictor models} 
\begin{figure*}[t]
    \includegraphics[width=\linewidth]{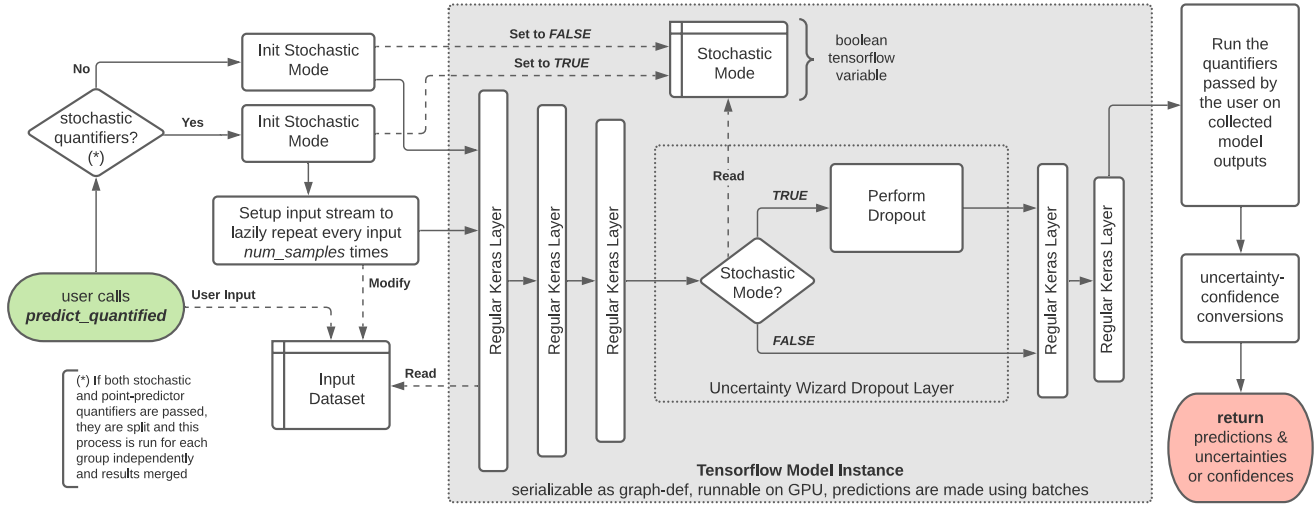}
  \caption{Process flow of a \lstinline{predict_quantified} call for stochastic and point-predictor models on an example model with five regular keras layer and one uncertainty wizard dropout layer.  }
  \label{fig:stochastic_model_prediction}
\end{figure*}

Given a model architecture that contains at least one dropout layer, 
we can use that model for both MC-Dropout and  Point Prediction uncertainty estimation, but while in the former case the dropout layers have to be enabled,
in the latter they have to be disabled.
Our models should hence be able to make quantifications based on both PPQ and SBQs, without having to keep two models.
Our solution to this requirement is to dynamically enable and disable randomized layers at prediction time, which we found to be more involved than first expected:
In plain \tfkeras, enabling and disabling happens based on a global \lstinline{learning_phase} state, which cannot be manipulated at prediction time for our purposes.
We solved this problem by implementing our own \tfkeras layers which wrap randomized \tfkeras layers. E.g., \lstinline{UwizBernoulliDropout} wraps \lstinline{keras.layers.Dropout.} 
This layer has access to a boolean state managed by \tool to which we refer as \emph{Stochastic Mode}:
This allows us to enable dropout if the model is being trained (\tfkeras default) \textit{or} if  Stochastic Mode is set to \lstinline{True}.
\tool sets the stochastic mode upon any call to \lstinline{predict_quantified}, setting its value to \lstinline{True} iff the passed quantifier is an SBQ. 
In that case, \tool also takes care of replicating the provided inputs the specified number of times, allowing to collect multiple samples.
For memory efficiency reasons, this is done in a \lstinline{tensorflow.data.Dataset} stream, allowing the broadcasting to be done only once needed.
This prediction process is illustrated in \autoref{fig:stochastic_model_prediction}.

To allow transparent but still highly flexible use, despite relying on using custom \tfkeras layer, we implemented our Sequential API in a way such that randomized \tfkeras layers (e.g. \lstinline{keras.layers.Dropout}) are automatically replaced by our wrapper layers, hiding the entire Stochastic Mode complexity from the user \exampleReferences{S5}. 
At the same time, to allow users to configure randomness to their need, our Functional API permits users to access and build upon the Stochastic Mode directly, allowing them not just to use our randomized layer implementations, but also to define custom, Stochastic Mode dependent layers.

\begin{listing}[t]
    \begin{minted}[fontsize=\scriptsize, linenos]{Python}
# Define how to create & train models
def supplier(model_id):
    model = keras.models.Sequential()
    model.add( ... )
    model.compile( ... )
    history = model.fit( ... )
    return model, history.history
# Create a lazy ensemble model instance
ensemble = uwiz.models.LazyEnsemble(
    path='an/empty/or/nonexistent/folder',
    num_models=20)
# Let uwiz create and train atomic models
train_histories = ensemble.create(
    supplier, num_processes=5)
# Calculate predictions & confidence
ensemble.predict_quantified(
    x_test, quantifiers='ensembling',
    num_processes=10)
    \end{minted}
    \caption{Lazy Ensemble Model}
    \label{lst:ex_lazy_ensemble}
\end{listing}

\subsection{Deep Ensembles}
Deep ensembles can be implemented naively by
looping over the training process for a single model the desired number of times and collecting the desired results.
The same can be done for model modification and prediction tasks.
However, there are two major practical performance disadvantages of such an approach:
\textit{(1) System Resources:} Large scale ensembles will eventually exceed the available system resources, in particular GPU dedicated memory and system RAM. 
    Memory leaks would further exacerbate this problem. 
\textit{(2) Low performance on small problems:} 
    In powerful workstations where modern GPUs are used to run ANNs, the available computing capacities might not be fully consumed by small models,
    making their execution suboptimal.
    This can easily go unnoticed as Tensorflow by default allocates all available memory on the GPU upon initialization.
    This problem is particularly evident in testing research, where most studies are based on datasets that only require small models, such as \emph{mnist} or \emph{cifar10}\cite{Riccio2020}. It becomes a major problem with deep ensembles, which require multiple, possibly small, models to be run at the same time.

Both problems described above can be solved by implementing countermeasures, such as persisting the models on disk, spawning child processes, and configuring them with  sensible GPU settings. 
This however requires to add a lot nontrivial and hard-to-maintain boilerplate code.

The main challenge behind implementing deep ensembles was thus to make it as simple as possible to run Deep Ensembles in a performance optimized way:
We want to allow the user to specify a single integer which defines the number $p$ of processes which should be used for model execution: When this is set to $0$, the model is executed in the main process, thus using the Tensorflow's default. Otherwise, new processes are created and the main process will not perform any model execution, preventing problems related to memory leaks.

The crucial component in our implementation to permit such a simple API is the \lstinline{EnsembleContextManager} (ECM) class.
Instances of such class  handle the actions delegated to every spawned process or evaluated in the main process.
Amongst others, they provide the functionality to save and load \tfkeras models, 
they initialize new processes and device configuration, e.g., by loading an appropriately configured Tensorflow environment,
and they define the number of atomic models for which each process is re-used, before being discarded and replaced with a freshly initialized process.

To provide sensible defaults, \tool implements three subclasses of ECM. 
They differ with respect to their tensorflow initialization  as follows:
\begin{description}[noitemsep]
\item [NoneContextManager]  uses Tensorflow's default settings without modification. Default when $p=0$.
\item [DynamicGpuGrowthContextManager] dynamically allocates GPU memory when needed, thus allowing multiple processes running on the same GPU. Default when $p>0$.
\item [DeviceAllocatorContextManager] offers
abstract methods, which have to be overridden by \tool's user. They allow to define how much GPU memory should be dedicated per atomic model, how many models can execute concurrently on a specific GPU and if a model should be executed on the CPU\footnote{Executing on a CPU is typically not recommended as it is generally slow and as the CPU resources may already be used for data pre-processing by the other processes}.
\end{description}

To allow maximum flexibility, \tool allows  advanced users to override ECMs and to pass their subclass when calling any method on the Deep Ensemble.

\subsection{Quantifiers}
\tool provides a selection of seven different quantifiers (2 PPQ, 5 SBQ), taken from the literature. 
They are implemented using only \textsc{numpy} operations,
thus allowing performance optimized quantification of large amounts of ANN outputs.
All quantifiers have aliases assigned (e.g. 'var\_ratio' for \lstinline{uwiz.quantifiers.VariationRatio}),
allowing users to easily refer to them by name when using them in \lstinline{predict_quantified} \exampleReferences{S11, S17, E17}.
Extension of the classes \lstinline{ConfidenceQuantifier} or \lstinline{UncertaintyQuantifier} further allows the users to define their own quantifiers.

\section{Assessment}
\label{sec:assessment}

We discuss the quality of \tool based on two aspects: 
First, we evaluate the performance of our Deep Ensemble implementation.
Then, we discuss code quality and user-friendliness -- the primary goal of \tool.

\subsection{Performance analysis of lazy ensembles}
To evaluate the performance of different ECMs, we trained a Deep Ensemble containing 20 atomic models for 100 epochs, using a batch size of 32 on the \textit{cifar10} dataset,
a popular dataset in machine learning based system testing~\cite{Riccio2020}.
\footnote{
Code:
\url{https://github.com/testingautomated-usi/repli-ensemble-bench}} 
We used the following hardware for our tests:
\begin{description}[noitemsep]
\item [R8] A high-end Alienware Aurora R8 gaming PC, running on Windows 10, with an Intel i7 9700 CPU and an RTX-2080Ti GPU.
\item [Cust.] A mid-range custom assembled PC running on Ubuntu 20.04 with a Threadripper 1920X CPU, and two GPUs: A GTX-1060 and a GTX-1070Ti. 
\end{description}
We deliberately ran a comparably high number of atomic models per GPU, which ensured that the GPU capacities were fully used.\footnote{Note that this does not imply an overall GPU load of 100\%, most likely due to capacity bottlenecks. For example for the 2080Ti, we observe that we cannot go above ~84\% GPU load. At that stage, the GPUs cuda load as reported by Windows Task Manager is at 100\%.}
Our results, using the different ECMs are shown in \autoref{tab:performance}. 
They indicate that we were able, through parallelization, to reduce the processing time on  Cust. by 67\% and on  R8 by 19\%.
Thus, \tool substantially helps to improve performance.
The results also support the following, practically relevant observation:
Despite being 45\% cheaper, the mid-range multi-GPU setting showed a 44\% better performance than the high-range single-GPU setting.\footnote{Prices from amazon.com, 19.12.2020; RTX-2080Ti: \$1450,  GTX-1060: \$300, GTX-1070Ti: \$500}

\begingroup
\setlength{\tabcolsep}{5.5pt} 
\renewcommand{\arraystretch}{1} 

\newcommand{\contexseparator}[0]{1ex}

\newcommand{\vertimodeltype}[1]{\begin{tabular}{@{}c@{}}\rotatebox[origin=c]{90}{\parbox{1cm}{\centering #1}}\end{tabular}}
\def \spacebetweenbigtrows {\vspace{0.15cm}}

\begin{table}[t]
\begin{tabular}{@{}llcccc@{}}
\toprule
PC                               & \begin{tabular}[l]{@{}l@{}}EnsembleContext- \\ Manager\end{tabular} & \begin{tabular}[c]{@{}c@{}}CPU \\ Load\end{tabular} & \begin{tabular}[c]{@{}c@{}}GPU0 Load\\ (processes)\end{tabular} & \begin{tabular}[c]{@{}c@{}}GPU1 Load\\ (processes)\end{tabular} & Time \\ \midrule
\multirow{2}{*}{R8}       & None (tf defaults)     & 14.7\%                                              & 46.9\% (1)                                                       & n.a.                                                             & 3:07h                                                        \\
 \spacebetweenbigtrows{}                                 & DynamicGrowth          & 50.3\%                                              & 83.9\% (5)                                                       & n.a.                                                             & \textbf{2:32h}                                                       \\
\multirow{3}{*}{Cust.} & None (tf defaults)     & 8.9\%                                               & 0.0\% (0)                                                        & 45.4\% (1)                                                       & 5:10h                                                        \\
                                 & DynamicGrowth          & 14.3\%                                              & 0.1\% (0)                                                        & 92.2\% (3)                                                       & 2:57h                                                        \\
                                 & DeviceAllocator & 26.8\%                                              & 98.2\% (3)                                                       & 92.4\% (3)                                                       & \textbf{1:41h}                                                        \\ \bottomrule
\end{tabular}
\vspace{0.1cm}
\caption{Performance gains using our ContextManagers}
\label{tab:performance}
\end{table}

\subsection{Code Quality and Usability}
In this section, we discuss the quality of \tool according to the criteria \emph{Documentation}, 
\emph{Testing} and \emph{Deployment}: 

\begin{description}[noitemsep]
\item [Documentation] Our documentation is publicly available on \textbf{\href{https://uncertainty-wizard.readthedocs.io}{uncertainty-wizard.readthedocs.io}}.
It consists of user guides, well suited to get started,
of examples of specific use-cases, which can be run and modified directly in google Colab, 
and of course the full API documentation:
\tool has a 100\% python Docstring coverage on its public modules, classes and functions.
\item [Testing] 
Our test suite consists of 114 unit tests which are all executed by our continuous integration (CI) for python versions 3.6, 3.7 and 3.8. Together, they lead to a test coverage of 91\%. 
The Jupyter examples of our documentation are smoke-tested for runtime errors by our CI as well.
\item [Deployment] \tool is open source (MIT license) and publicly hosted on github.
In addition, it is registered in the \emph{python package index (\href{https://pypi.org}{pypi.org})} allowing easy installation using \lstinline{pip install uncertainty-wizard}. 
Acknowledging the high importance of a slim dependency tree, the only dependency is a recent version of Tensorflow. 

\end{description}

We say our approach is near-transparent because the code changes required for its integration are minimal. For a sequential model, only two lines of code have to be changed to make it stochastic: the constructor and the prediction call. Training an ensemble (using a numpy dataset) requires the refactoring of the creation and training function into a serializable function.
Then, only three new statements have to be added:
the ensemble constructor, 
the create call, and a prediction call.
\section{Conclusion}
\label{sec:conclusion}

With MC-Dropout and Deep Ensembles, the machine learning literature provides two ways to quantify uncertainty, which add to Point Predictors. 
These approaches have already been used to generate/prioritize test data and to supervise/heal neural networks.
However, their implementation from scratch requires a large amount of hard-to-maintain boilerplate code.
\tool builds on the  popular machine learning framework \emph{Tensorflow} and its easy to use interface \tfkeras, implementing the uncertainty quantifications: Point Predictors, MC-Dropout and Deep Ensembles.
Users can easily, near transparently and highly efficiently identify the ANN inputs which are most (or least) likely to cause wrong ANN predictions.

\typeout{}
\bibliographystyle{IEEEtran}
\bibliography{main}

\begin{thebibliography}{10}
\providecommand{\url}[1]{#1}
\csname url@samestyle\endcsname
\providecommand{\newblock}{\relax}
\providecommand{\bibinfo}[2]{#2}
\providecommand{\BIBentrySTDinterwordspacing}{\spaceskip=0pt\relax}
\providecommand{\BIBentryALTinterwordstretchfactor}{4}
\providecommand{\BIBentryALTinterwordspacing}{\spaceskip=\fontdimen2\font plus
\BIBentryALTinterwordstretchfactor\fontdimen3\font minus
  \fontdimen4\font\relax}
\providecommand{\BIBforeignlanguage}[2]{{%
\expandafter\ifx\csname l@#1\endcsname\relax
\typeout{** WARNING: IEEEtran.bst: No hyphenation pattern has been}%
\typeout{** loaded for the language `#1'. Using the pattern for}%
\typeout{** the default language instead.}%
\else
\language=\csname l@#1\endcsname
\fi
#2}}
\providecommand{\BIBdecl}{\relax}
\BIBdecl

\bibitem{Weiss2021}
M.~Weiss and P.~Tonella, ``Fail-safe execution of deep learning based systems
  through uncertainty monitoring,'' in \emph{2021 IEEE 14th International
  Conference on Software Testing, Validation and Verification (ICST)}.\hskip
  1em plus 0.5em minus 0.4em\relax IEEE, 2021, forthcoming.

\bibitem{Kim2018}
J.~Kim, R.~Feldt, and S.~Yoo, ``Guiding deep learning system testing using
  surprise adequacy,'' in \emph{2019 IEEE/ACM 41st International Conference on
  Software Engineering (ICSE)}.\hskip 1em plus 0.5em minus 0.4em\relax IEEE,
  2019, pp. 1039--1049.

\bibitem{Wang2020}
H.~Wang, J.~Xu, C.~Xu, X.~Ma, and J.~Lu, ``Dissector: Input validation for deep
  learning applications by crossing-layer dissection,'' in \emph{Proceedings of
  42nd International Conference on Software Engineering}.\hskip 1em plus 0.5em
  minus 0.4em\relax ACM, 2020.

\bibitem{Stocco2020}
A.~Stocco, M.~Weiss, M.~Calzana, and P.~Tonella, ``Misbehaviour prediction for
  autonomous driving systems,'' in \emph{Proceedings of 42nd International
  Conference on Software Engineering}.\hskip 1em plus 0.5em minus 0.4em\relax
  ACM, 2020, p. 12 pages.

\bibitem{Zhang2020}
X.~Zhang, X.~Xie, L.~Ma, X.~Du, Q.~Hu, Y.~Liu, J.~Zhao, and M.~Sun, ``Towards
  characterizing adversarial defects of deep learning software from the lens of
  uncertainty,'' in \emph{Proceedings of 42nd International Conference on
  Software Engineering}.\hskip 1em plus 0.5em minus 0.4em\relax ACM, 2020.

\bibitem{Berend2020}
D.~Berend, X.~Xie, L.~Ma, L.~Zhou, Y.~Liu, C.~Xu, and J.~Zhao, ``Cats are not
  fish: Deep learning testing calls for out-of-distribution awareness,'' in
  \emph{The 35th IEEE/ACM International Conference on Automated Software
  Engineering}.\hskip 1em plus 0.5em minus 0.4em\relax New York, NY, USA:
  Association for Computing Machinery, 2020.

\bibitem{Jospin2020}
L.~V. Jospin, W.~Buntine, F.~Boussaid, H.~Laga, and M.~Bennamoun, ``Hands-on
  bayesian neural networks -- a tutorial for deep learning users,'' 2020.

\bibitem{Gal2016}
Y.~Gal and Z.~Ghahramani, ``Dropout as a bayesian approximation: Representing
  model uncertainty in deep learning,'' in \emph{Proceedings of the 33rd
  International Conference on International Conference on Machine Learning -
  Volume 48}, ser. ICML'16.\hskip 1em plus 0.5em minus 0.4em\relax JMLR.org,
  2016, pp. 1050--1059.

\bibitem{Lakshminarayanan2017}
B.~Lakshminarayanan, A.~Pritzel, and C.~Blundell, ``Simple and scalable
  predictive uncertainty estimation using deep ensembles,'' in \emph{Advances
  in neural information processing systems}, 2017, pp. 6402--6413.

\bibitem{Ovadia2019}
Y.~Ovadia, E.~Fertig, J.~Ren, Z.~Nado, D.~Sculley, S.~Nowozin, J.~Dillon,
  B.~Lakshminarayanan, and J.~Snoek, ``Can you trust your models uncertainty?
  evaluating predictive uncertainty under dataset shift,'' \emph{Advances in
  Neural Information Processing Systems}, pp. 13\,991--14\,002, 2019.

\bibitem{Riccio2020}
V.~Riccio, G.~Jahangiroba, A.~Stocco, N.~Humbatova, M.~Weiss, and P.~Tonella,
  ``Testing machine learning based systems: a systematic mapping,''
  \emph{Empirical Software Engineering}, 2020.

\end{thebibliography}

\end{document}